\pdfoutput=1
\documentclass[letterpaper, 10pt, conference]{ieeeconf}      
\IEEEoverridecommandlockouts                              
\usepackage{algorithm}
\usepackage{multicol}
\usepackage{multirow}
\usepackage{algpseudocode}
\usepackage{color}
\usepackage{xcolor}
\usepackage{graphicx, epstopdf}
\usepackage{bm}
\usepackage{amsmath}
\usepackage{amssymb}
\usepackage{tabularx}
\usepackage{subcaption}
\usepackage{url}
\usepackage{cite}
\usepackage{prettyref}
\usepackage{booktabs}
\usepackage{siunitx}

\newrefformat{fig}{Figure~\ref{#1}}
\newrefformat{sec}{Section~\ref{#1}}
\newrefformat{tab}{Table~\ref{#1}}
\newrefformat{alg}{Algorithm~\ref{#1}}
\newrefformat{eqn}{Equation~\ref{#1}}

\usepackage{hyperref}
\hypersetup{
    colorlinks,
    linkcolor={blue!80!black},
    citecolor={blue!80!black},
    urlcolor={blue!100!black}
}

\makeatletter
\newcommand{\printfnsymbol}[1]{%
  \textsuperscript{\@fnsymbol{#1}}%
}
\makeatother

\title{
\LARGE \bf 
DynamicFilter: an Online Dynamic Objects Removal Framework for Highly Dynamic Environments
} %Use for final RAL version 

\author{Tingxiang Fan*, Bowen Shen*, Hua Chen, Wei Zhang$^{\dagger}$ and Jia Pan$^{\dagger}$%
\thanks{* denotes equal contribution, $\dagger$ denotes corresponding author.}
\thanks{T. Fan, J. Pan are with the University of Hong Kong. B. Shen, H. Chen, W. Zhang are with the Southern University of Science and Technology.}

\thanks{This work is supported in part by HKSAR 
Research Grants Council (RGC) General Research Fund (GRF) HKU 11202119, 11207818, the Innovation and Technology Commission of the HKSAR Government under the InnoHK initiative, the National Natural Science Foundation of China under Grants 62073159 and 62003155, the Shenzhen Science and Technology Program under Grant JCYJ20200109141601708, and the Science, Technology and Innovation Commission of Shenzhen Municipality under grant ZDSYS20200811143601004}
}
\begin{document}
\maketitle
%\thispagestyle{empty}
%\pagestyle{empty}

%%%%%%%%%%%%%%%%%%%%%%%%%%%%%%%%%%%%%%%%%%%%%%%%%%%%%%%%%%%%%%%%%%%%%%%%%%%%%%%%

\begin{abstract}

Emergence of massive dynamic objects will diversify spatial structures when robots navigate in urban environments. Therefore, the online removal of dynamic objects is critical. In this paper, we introduce a novel online removal framework for highly dynamic urban environments. The framework consists of the scan-to-map front-end and the map-to-map back-end modules. Both the front- and back-ends deeply integrate the visibility-based approach and map-based approach. The experiments validate the framework in highly dynamic simulation scenarios and real-world dataset.

\end{abstract}

\section{Introduction}
\label{sec:intro}

% autonomous driving need HD Map, the beneath issue is scene understanding. Online scene understanding is the key. 

% With the intensive development on autonomous driving technologies in recent years, robots are moving to the human-live urban environment from structured industrial scenarios. 
Long-range navigation in urban environments is a challenging problem for modern autonomous robotic systems. The High-Definition Map (HD Map) that can provide a highly accurate representation of the urban environment has been widely used for scene understanding and planning. However, constructing and maintaining the HD Map is time-consuming and laborious as it needs to operate a large number of vehicles for data collection to continuously update the map. By online understanding the surrounding scenarios, humans can navigate without the prior assistance of HD Map. Therefore, online scene understanding is a promising direction to eliminate robots' dependence on HD Maps.

% Online dynamic object removal is still challenged. 

By leveraging 3D LiDAR sensors, robots can perform simultaneous localization and mapping (SLAM) to online reconstruct the spatial structure of surrounding environments~\cite{shan2020lio,xu2021fast}. However, most existing SLAM techniques assume the perceived objects are static and time-invariant. When a robot navigates in urban environments, the emergence of massive dynamic objects will diversify the spatial structure. Therefore, the online removal of dynamic objects is critical.

\begin{figure}
\centering
\begin{subfigure}{0.45\textwidth}
\centering
\includegraphics[width=1\linewidth]{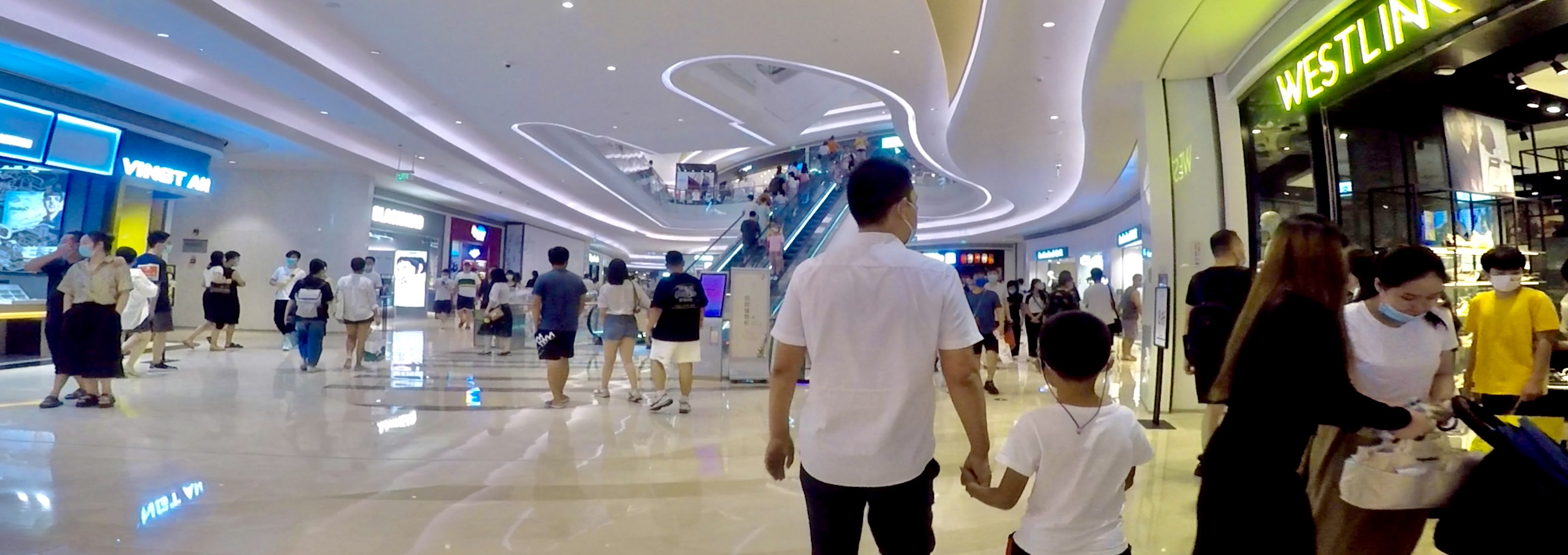}
% \caption{}
\vspace{-0.15in}
\label{fig:mall}
\end{subfigure} \\
\begin{subfigure}{0.45\textwidth}
\centering
\includegraphics[width=1\linewidth]{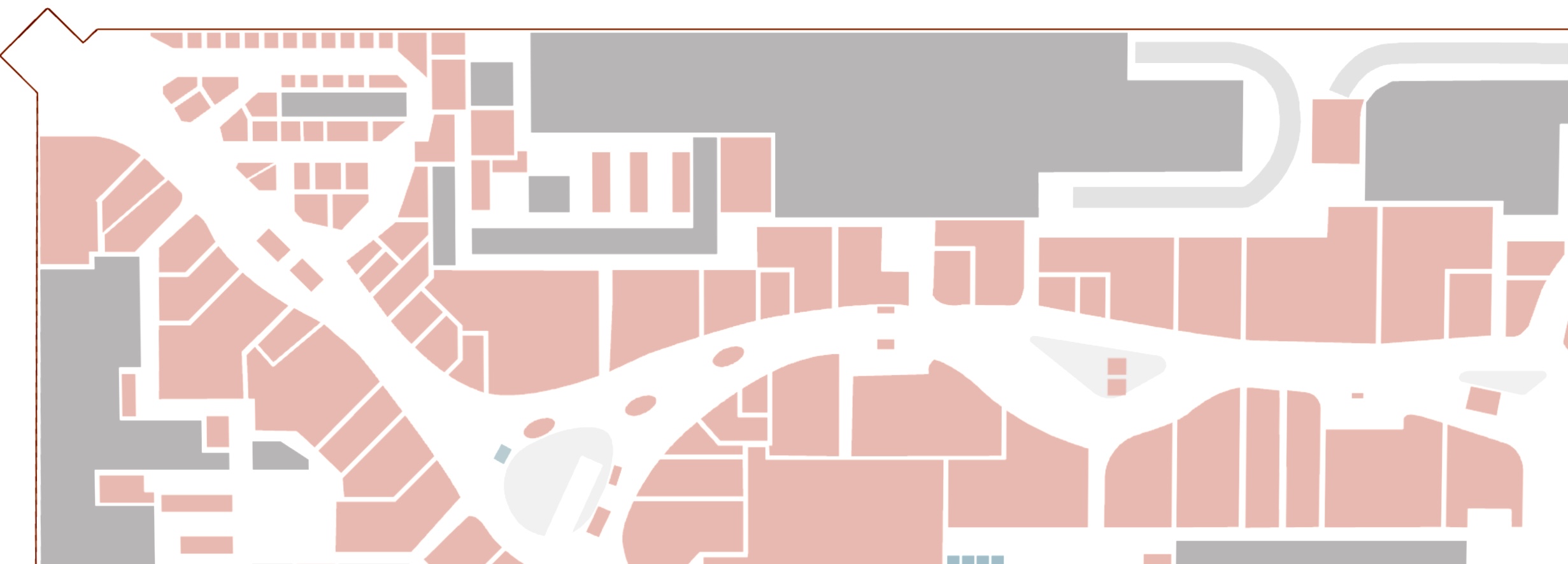}
% \caption{The floor plan of the mall.}
\vspace{-0.1in}
\label{fig:mall_layout}
\end{subfigure} \\
\begin{subfigure}{0.45\textwidth}
\centering
\includegraphics[width=1\linewidth]{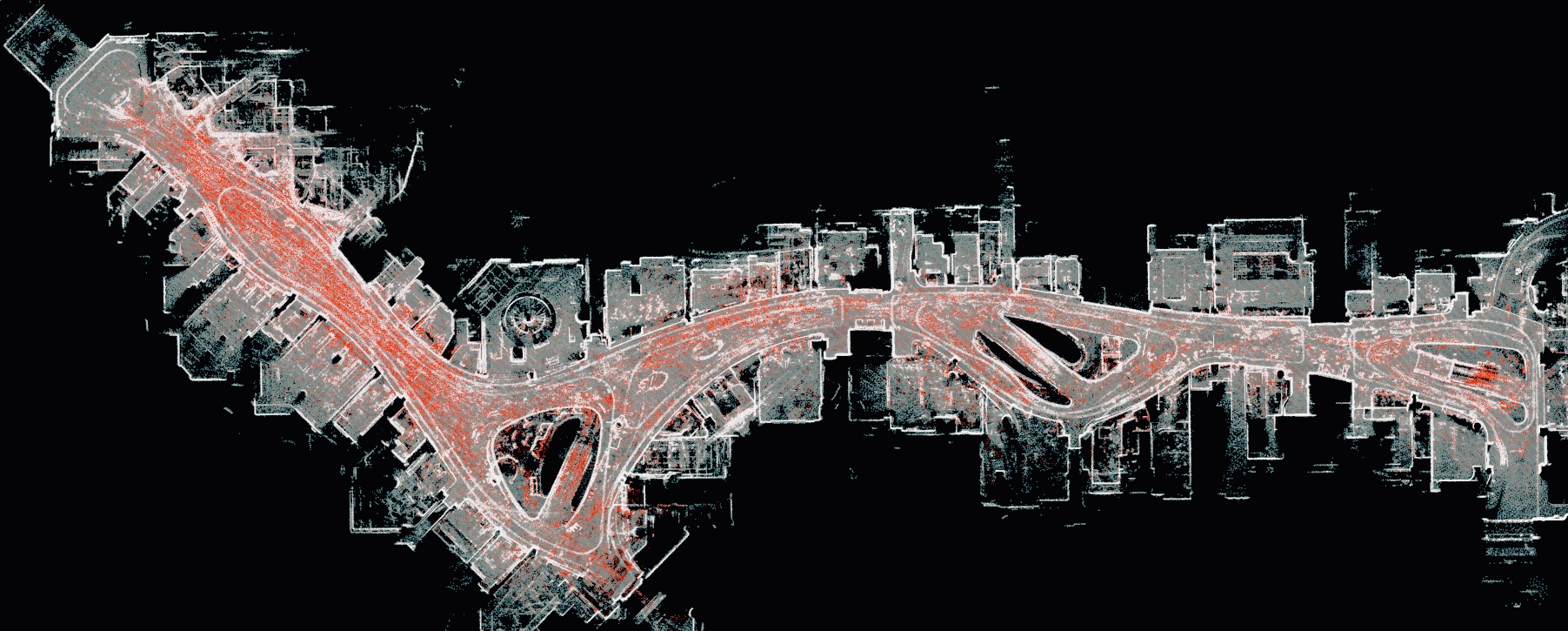}
% \caption{The point cloud map built by our approach. The red points are the dynamic pedestrians are removed. The white points are the spatial structure of the environment.}
\vspace{-0.15in}
\label{fig:mall_dynamic}
\end{subfigure} 
\caption{\textit{Top}: Collecting data with dense crowds in the mall. \textit{Medium}: The floor plan of the mall. \textit{Bottom}: The point cloud map built by our approach. The red points are the dynamic pedestrians to be removed. The white points are the spatial structure of the environment.}
\label{fig:cover_image}
\vspace{-0.3in}
\end{figure}

% Map-based and Visibility-based exist their shortcomings. 
In this paper, we focus on the detection and removal of the dynamic points perceived by the 3D LiDAR. It is challenging to detect dynamic points from a single scan. Although learning-based approaches can instantly predict the dynamic segments from a single scan~\cite{cortinhal2020salsanext, milioto2019rangenet++}, they rely on numerous labeled datasets and may fail when encountering unlabeled classes. By considering the inconsistency between multiple scans, the occupancy map-based and visibility-based approaches can estimate the dynamic points without any semantics labels. However, both approaches have their own shortcomings. The occupancy map-based method~\cite{hornung2013octomap,schauer2018peopleremover,pfreundschuh2021dynamic}, leveraging the ray-tracing, can estimate the occupancy probability in the grid space. Although it has been widely used in 2D mapping, the computational cost of real-time ray-tracing for 3D points is unaffordable. To alleviate the computational burden, the visibility-based method~\cite{pomerleau2014long,yoon2019mapless,kim2020remove} re-projects the range measurements of the 3D point cloud to the image plane according to the specific field of view (FOV) and resolution. The visibility difference between multiple scans at each pixel is considered as dynamic points. However, it suffers from \textit{incidence angle ambiguity}\cite{pomerleau2014long, wurm2010octomap} and \textit{occlusion}\cite{lim2021erasor} issues. These \textit{visibility issues} will decrease its performance.

\begin{figure*}
\centering
\vspace*{-0.0in}
\includegraphics[width=0.9\linewidth]{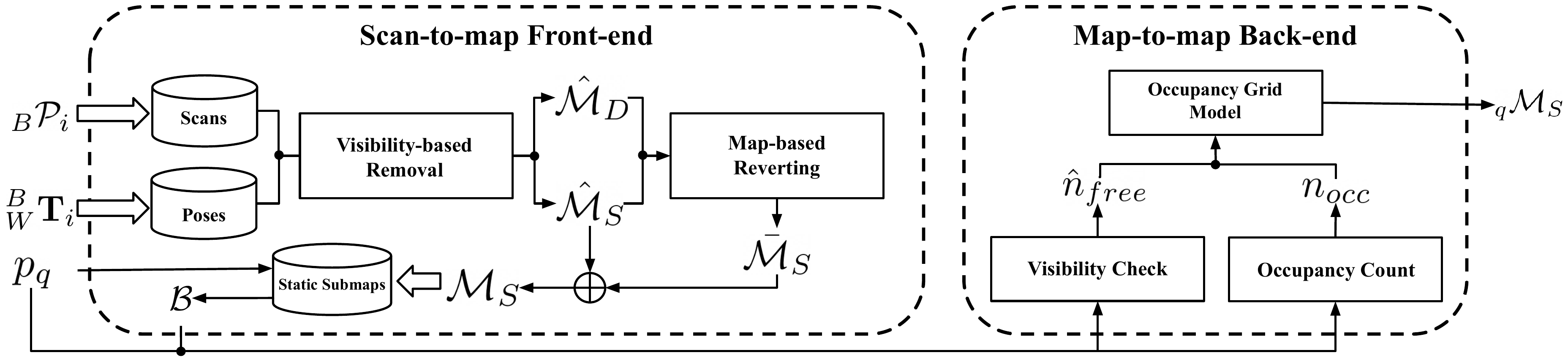}
\caption{Our pipeline is composed of front-end and back-end executed in parallel, with the visibility-based and map-based methods integrated.}
\label{fig:pipeline}
\vspace*{-0.25in}
\end{figure*}

% Our tightly-coupled method can address above issues. 
% scan-to-map map-to-map 

In this paper, we propose an online dynamic object removal framework that integrates both the map-based and the visibility-based methods. Inspired by the modern SLAM paradigm~\cite{bresson2017slam_survey}, our framework is divided into the scan-to-map front-end and the map-to-map back-end. The front-end is time-critical that needs to instantly compute coarse results, while the back-end could be computation-intensive to provide further optimization results with historical data. Specifically, our framework takes consecutive scans and their poses as input. First, the visibility-based approach is used for dynamic object removal given its superior computational efficiency. After the removal, the raw static submap $\hat{\mathcal{M}}_{S}$ can be obtained. Note that $\hat{\mathcal{M}}_{S}$ preserves sparse static points due to the \textit{visibility issues} mentioned above. Then, we propose the map-based reverting approach to recover denser static points $\bar{\mathcal{M}}_S$ from the sparse features. Due to the limited horizon of consecutive frames, the static submap $\mathcal{M}_S$ generated by the front-end may not be satisfactory. Finally, the map-to-map back-end is introduced to further optimize the result. The core idea of the back-end is to compute the occupancy probability in each voxel from multiple submaps. The voxel is marked as free when a ray passes through the voxel, where we use \textit{visibility check} to approximately calculate $\hat{n}_{\text{free}}$, the number of rays passing through each voxel, for accelerating the ray-tracing process. The entire pipeline is demonstrated in \prettyref{fig:pipeline} and the main contributions are summarized as follows:
\begin{itemize}
    \item We propose the online dynamic object removal framework that fuses the visibility-based and map-based approaches. To our best knowledge, it is the first work proposing the front-end and back-end structures for dynamic object removal. 
    
    \item We present the \textit{visibility check} that uses the visibility-based approach to approximate the ray-tracing process and accelerate the occupancy computation. 
    
    \item We propose the map-based reverting algorithm in the front-end which can mitigate the \textit{visibility issues} caused by the visibility-based removal. 
    
    \item We quantitatively evaluate the algorithm's performance in complex simulation scenarios and qualitatively demonstrate the online performance in a diverse real-world scenes that are crowded. 
\end{itemize}

% \vspace{-0.05in}
\section{Related Work}
\label{sec:related_work}
\vspace{-0.05in}
In this section, we review the related work about dynamic object removal from 3D point clouds, which can be roughly divided into the following three categories:

% learning-based
\textbf{Model-dependent approaches.} Dynamic points can be detected based on the prior model. Some methods rely on the ground plane model to separate dynamic objects from the ground. \cite{dewan2016motion, dewan2016rigid} require a pre-processing step of ground removal. \cite{lim2021erasor} is based on the premise that all dynamic objects will be in contact with the ground, so all points except the ground can be filtered out where dynamic objects may exist. Therefore, these methods' performance may degrade in situations where such assumption does not hold. 

More generally, with recent advances in deep learning, some dynamic objects such as pedestrians and vehicles can be detected according to their appearance models~\cite{cortinhal2020salsanext, milioto2019rangenet++}. However, the detected results only indicate that the objects are potentially dynamic but do not mean that the objects actually are moving. For example, for a car parked in a parking lot, the robot should understand this object as a static obstacle instead of a dynamic object. \cite{chen2021lidar-mos} takes the sequential range images as input to capture the dynamic points based on the inconsistency between these images. However, these learning-based approaches rely on manually labeled high-quality datasets.

% occupancy-based
\textbf{Occupancy map-based approaches.} By tracking the emitted ray from LiDAR, the robot can consider the ray end point as being occupied and the voxels that the ray traverses as free. In this way, the occupancy probability in the surrounding environment can be computed. However, ray-tracing for 3D LiDAR is computationally expensive. Even with some engineering optimization~\cite{hornung2013octomap}, it is still challenging to process massive 3D data online, though it is indeed an effective method to filter out dynamic points~\cite{schauer2018peopleremover}. \cite{pfreundschuh2021dynamic} introduces the occupancy map-based method for offline labeling of dynamic objects and the labeling results can be used as the ground truth for the training step of learning-based approaches.

% visibility-based
\textbf{Visibility-based approaches.} Compared to the occupancy map-based methods, the visibility-based approaches only compute the visibility difference~\cite{pomerleau2014long,yoon2019mapless,kim2020remove}. Specifically, if the observed point is occluded in the line of sight of the previously observed point, this point should be regarded as dynamic. It dramatically reduces the computational cost compared to ray-tracing, though such simplification is also accompanied with performance loss. In the situation of a large incident angle~\cite{pomerleau2014long, wurm2010octomap}, or when occlusion occurs~\cite{kim2020remove,lim2021erasor}, the visibility-based method may not be able to filter out the dynamic points correctly (Fig. 2 in \cite{lim2021erasor}).  \cite{kim2020remove} proposes the \textit{removing-and-reverting} mechanism by iteratively retaining the static points from falsely removed points. However, the reverting process is still based on the visibility process and thus its performance is still limited due to the \textit{visibility issues}. Moreover, \cite{pomerleau2014long, kim2020remove} only implement their approaches under the offline paradigm. 

In this paper, we aim to develop a model-independent approach that does not utilize any existing model. By fusing occupancy map-based and visibility-based methods, our proposed method is dedicated to the online removal of dynamic points with high efficiency and high accuracy. 

% TODO: online vs offline,

% \input{problem}
% \vspace{-0.05in}
\section{Methodology}
\label{sec:methodology}
\vspace{-0.05in}
\begin{figure*}
\centering
\vspace*{-0.0in}
\includegraphics[width=1.0\linewidth]{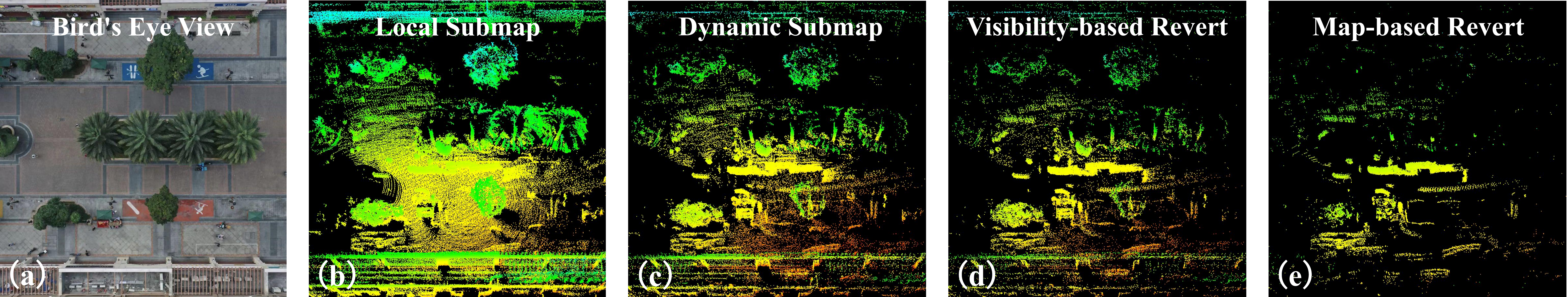}
\caption{The qualitative comparison between visibility-based reverting~\cite{kim2020remove} and our map-based reverting. \textit{(a)} is the bird's eye view of the current scene. By stacking the consecutive scans, we can acquire the raw local submap (i.e. \textit{(b)}) in this scene. The point cloud is colored according to its height. After the visibility-based removal, the detected dynamic points $\hat{\mathcal{M}}_D$ are visualized in \textit{(c)}. With visibility-based reverting, some falsely detected dynamic points are reverted to the static map as shown in \textit{(d)}. We present the result of our proposed map-based reverting in \textit{(e)}, which dramatically recovers the falsely detected dynamic points without losing dynamic features.}
\label{fig:revert}
\vspace*{-0.2in}
\end{figure*}

The online dynamic object removal framework consists of two parts: the scan-to-map front-end and the map-to-map back-end, as shown in \prettyref{fig:pipeline}. 
 
% Note that the front-end and back-end are executed in parallel. 

% The front-end is responsible for constructing a submap from the consecutive frames in real-time. At this stage, an efficient visibility-based method~\cite{kim2020remove} is used to roughly remove dynamic points in the submap. Then, the pre-processed submap is added into the submap buffer for the back-end to compute the occupancy probability from multiple submaps. 
% Note that, the front-end and back-end are executed in parallel. 

\vspace*{-0.05in}
\subsection{Problem setup}
We focus on the removal of dynamic points from the consecutive 3D LiDAR scans. The point cloud $_B\mathcal{P}_i$ of each scan locates in the local sensor frame $B$. By using state-of-the-art LiDAR-based SLAM approaches such as~\cite{shan2020lio,xu2021fast}, we can compute the transition $^B_W\mathbf{T}_i$ between local sensor frame $B$ and the global frame $W$.
% Thus, we associate the point cloud and its transition as a frame $_B\mathcal{F}_i$.

\vspace*{-0.05in}
\subsection{Scan-to-Map front-end}
The front-end takes $n$ temporally consecutive frames as input, then transforms these points into a global coordinate, and finally constructs an unprocessed submap
$\{ _W\mathcal{M} \mid \sum_{i}^{n}  {}^B_W\mathbf{T}_i \times {}_B\mathcal{P}_i \} $.

\subsubsection{Visibility-based removal}
\label{sec:visibility_removal}
The basic removal process is proposed by ~\cite{pomerleau2014long, kim2020remove}, which can be briefly summarized as follows. First, we project each frame of point cloud $_B\mathcal{P}_i$ and the unprocessed submap $_W\mathcal{M}$ to the image plane in the local coordinate system. Then, their range images $_{B}I^{\mathcal{P}}$ and $_{B}I^{\mathcal{M}}$ can be computed respectively. By comparing the difference between these two range images, we can detect the dynamic points as the submap $\hat{\mathcal{M}}_{D_i}$. After processing all frames in the submap, we can roughly divide the processed points into two classes: the static submap $\hat{\mathcal{M}}_{S}$ that only includes static points, and the dynamic submap $\hat{\mathcal{M}}_{D}$ that only includes the dynamic points.

\subsubsection{Map-based reverting}
However, the performance of the visibility-based removal will degrade in cases with large incident angles and occlusions. The manifestation of such degradation is that many static points are falsely filtered out as dynamic points. Inspired by ~\cite{kim2020remove}, our scan-to-map front-end also adopts the reverting phase to recover falsely detected static points. \cite{kim2020remove} repeatedly executes visibility-based removal but takes the dynamic submap $\hat{\mathcal{M}}_{D}$ as input, thereby continuously recovering the static points from the dynamic submap $\hat{\mathcal{M}}_{D}$. However, this reverting method may still suffer from the \textit{visibility issues}. 

Unlike~\cite{kim2020remove}, we introduce the map-based reverting to alleviate the \textit{visibility issues}. The insight behind is that the static submap $\hat{\mathcal{M}}_{S}$ is sparse and conservative, which strictly preserves the static points. In terms of the static points that were falsely removed, some work~\cite{kim2020remove,lim2021erasor} indicate that they often occur on the ground (caused by large incident angle) or at narrow objects, and boundary of objects (caused by occlusion). 

Therefore, we revert static points based on the set distance between the removed dynamic points and the static submap in the global coordinate. In particular, for each dynamic point in the dynamic map, we first search for its nearest points in the static map. Then, we employ the principal component analysis (PCA)~\cite{wold1987principal} to compute the eigen vector of the set of nearest points. Similarly, another eigen vector can also be computed for the union of this set of nearest points and the dynamic point itself. Finally, by comparing the normal distance between these two eigen vectors, we can get the distribution change after adding the dynamic point. If the change is small enough, we revert the dynamic point $\hat{\mathcal{M}}_D$ as static $\bar{\mathcal{M}}_S$. 
% The overall algorithm is described in \prettyref{alg:revert}, with 
The qualitative result is visualized in \prettyref{fig:revert}. Note that the reverting process does not rely on any model/plane assumption and thus is model-independent.

% \begin{algorithm}
% \caption{Map-based Reverting}
% \label{alg:revert}
% \begin{algorithmic}[1]
% \State \textbf{Input}: the static submap $\hat{\mathcal{M}}_S$, the dynamic submap $\hat{\mathcal{M}}_D$, search radius $r$ and distance threshold $dist_{\max}$.
% \State $\mathcal{D}$ = \{\}
% \State // \textit{Build kd-tree for nearest searching}
% \State $\mathcal{T}$ = kd-tree($\hat{\mathcal{M}}_S$)
% \For {$p_i$ \textbf{in} $\hat{\mathcal{M}}_D $}
%     % \State // \textit{Search the nearest points in the static submap}
%     \State $\mathcal{P}_i$ = $\mathcal{T}$.searchRadius($p_i$, $r$).
%     \State // \textit{Eigen vector in the static submap}
%     \State $\vec{\mathbf{v}}_S$ = PCA($\mathcal{P}_i$)
%     \State $\mathcal{P}_i'$ = $\{\mathcal{P}_i, p_i\}$
%     \State $\vec{\mathbf{v}}_{\text{new}}$ = PCA($\mathcal{P}_i'$)
%     \State // \textit{distribution distance after adding $p_i$}
%     \State $dist$ = ($\vec{\mathbf{v}}_{\text{new}} - \vec{\mathbf{v}}_S$).norm()
%     \State // \textit{revert the point close to the static map}
%     \If{$dist < dist_{\max}$}
%         \State $\mathcal{D}$.append($p_i$)
%     \EndIf
% \EndFor
% \State \textbf{Output}: the reverting point set $\mathcal{D}$.

% \end{algorithmic}
% \end{algorithm}

After map-based reverting, we obtain a static submap $\mathcal{M}_S$ and add it in the submap buffer along with its relative pose to the global frame averaged over the entire sequence.

\vspace*{-0.05in}
\subsection{Map-to-map back-end}
The front-end has the potential to remove all dynamic points in the long sequence~\cite{pomerleau2014long,kim2020remove} under the offline paradigm. However, given the online nature of our work, only short sequences are available, which will limit the front-end performance and thus the front-end may be only effective for some fast-moving objects. Therefore, we propose the map-to-map back-end to continuously improve the removal performance.  

\subsubsection{Occupancy grid model}
Given a query pose $p_q$, we can search for the nearest submaps in the submap buffer within a specific radius. The core idea is to combine multiple nearest static submaps to estimate the occupancy grid map in the environment using the occupancy grid model, which is simple but effective. Beforehand, we discretize the nearby space, i.e., finishing the voxelization step. We then count the number of times each voxel in the space is occupied (i.e., $n_{\text{occ}}$), and the number of times each voxel is passed through (i.e., $n_{\text{free}}$). The occupancy probability is then computed as $\frac{n_{\text{occ}}}{n_{\text{occ}} + n_{\text{free}}}$. $n_{\text{occ}}$ can be directly counted given the point cloud, but $n_{\text{free}}$ requires the ray-tracing computation. However, the online ray-tracing for each submap is intractable. Thus, we introduce the \textit{visibility check} to approximately estimate the denominator of the occupancy probability (i.e. $n_{\text{check}} = (n_{\text{occ}} + \hat{n}_{\text{free}})$).

\begin{figure}
\centering
\begin{subfigure}{0.5\textwidth}
\centering
\includegraphics[width=0.8\linewidth]{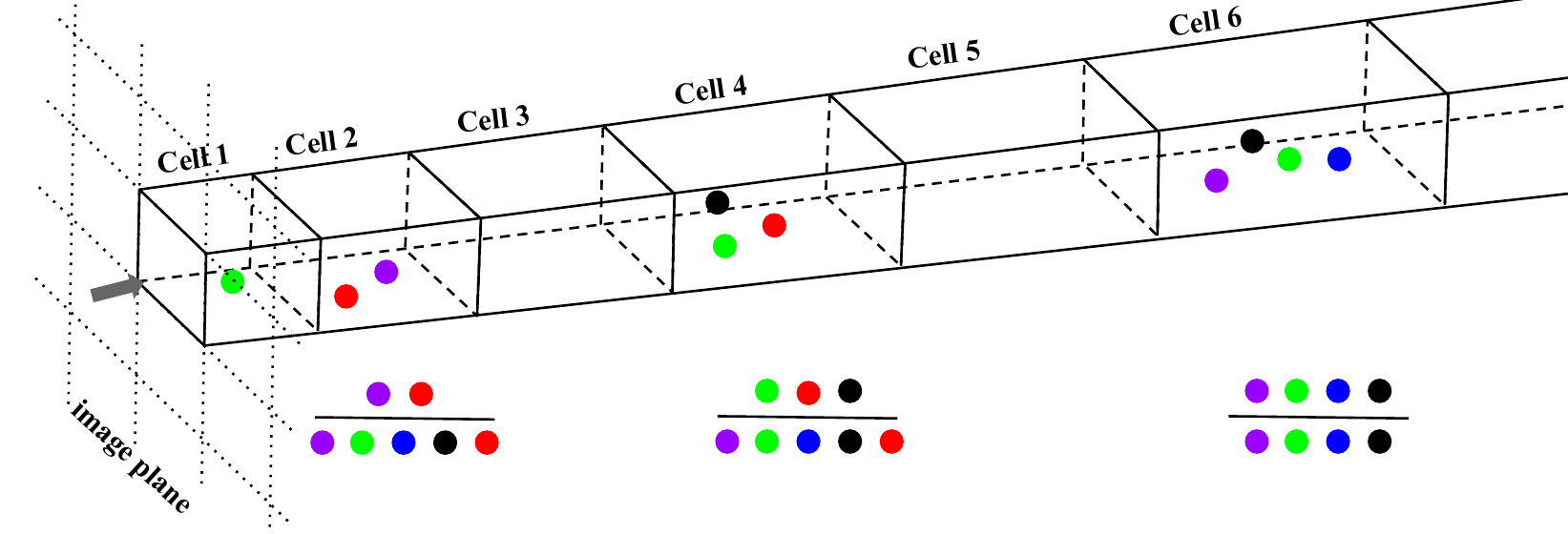}
\caption{Different colors represent points from different submaps and the grey arrow denotes the direction of the line of sight. Four submaps can reach the farthest cell (i.e., cell $6$) in the line of sight, so the occupied $n_{\text{occ}}$ is $4$ and the estimated free number$\hat{n}_{\text{free}}$ is $0$. Its occupancy probability is $\frac{4}{4}$. 
The cells before cell $6$ tend to be free in cases without occlusion. Therefore, for the cell $4$ whose $n_{\text{occ}}$ is $3$, it may be passed through by five submaps, and the occupancy probability is $\frac{3}{5}$. Similarly, the occupancy probability of cell $2$ is $\frac{2}{5}$.}

\label{fig:broadcast}
\end{subfigure} \\
\begin{subfigure}{0.5\textwidth}
\centering
\includegraphics[width=1.0\linewidth]{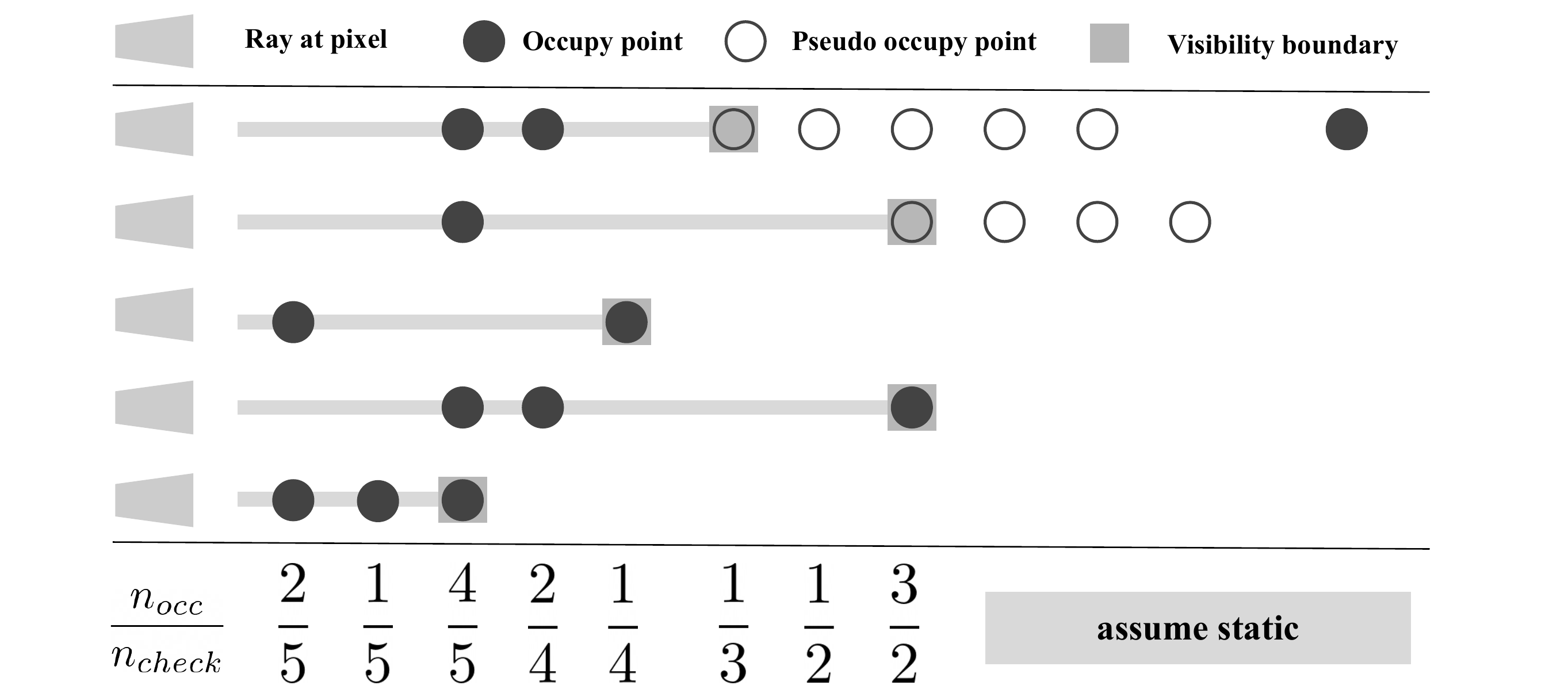}
\caption{One example of computing the occupancy probability. Different rows represent the rays for different submaps at a same pixel. After distinguishing the occupied point and pseudo occupied point, the \textit{visibility boundary} can be determined. We count the number of times each cell falling in a \textit{visibility boundary} as the $n_{\text{check}}$. Then, the occupancy probability can be computed.}
\label{fig:occ_prob}
\end{subfigure} 
\caption{The ray-tracing approximation by \textit{visibility check}. }
\label{fig:visibility_check}
\vspace{-0.25in}
\end{figure}

\begin{algorithm}[h]
\caption{Visibility check}
\label{alg:visibility_check}
\begin{algorithmic}[1]
\State \textbf{Input}:  nearest static submaps buffer $\mathcal{B}$, query pose $\mathbf{p}_q$, range depth $depth$, image resolution $res$, image FOV $fov$, and incident angle threshold $\lambda_{\text{thres}}$.
% \State // \textit{search nearest submaps in $\mathcal{B}$}
% \State $\mathcal{B}_q$ = $\mathcal{B}$.searchRadius($\mathbf{p}_q$, $r$)
\State $\mathcal{D}_{I}$ = \{\}
\For{${\mathcal{M}_S}$ \textbf{in} $\mathcal{B}$}
    \State // \textit{Transform submap using the query pose $\mathbf{p}_q$}
    \State ${_q\mathcal{M}_S}$ = transformMap(${\mathcal{M}_S}$, $p_q$)
    \State // \textit{Compute normal vectors for each submap}
    \State ${_q\mathcal{N}}$ = computeNormal(${_q\mathcal{M}_S}$)
    \State // \textit{Initialize image with ($inf$)}
    \State $I_{\text{bound}}$ = initializeImage($res$, $fov$, $inf$)
    \State $I_{\max}$ = initializeImage($res$, $fov$, $0$)
    \For {$i = 1,...,$ ${_q\mathcal{M}_S^i}$.size() }
        \State $row$, $col$ = reprojectToImage(${_q\mathcal{M}_S}[i]$)
        % \State // \textit{Compute range}
        \State $r$ = computeRange(${_q\mathcal{M}_S}[i]$)
        % \State // \textit{Compute incident angle}
        \State $\lambda$ = computeIncidentAngle(${_q\mathcal{N}}[i]$)
        \If{$\lambda > \lambda_{\text{thres}}$}
            \If{$r < I_{\text{bound}}$[$row$, $col$]}
                \State $I_{\text{bound}}$[$row$, $col$] = $r$
                \State $I_{\max}$[$row$, $col$] = $r$
            \EndIf
        \EndIf
    \EndFor

    \For {$i = 1,...,$ ${_q\mathcal{M}_S^i}$.size()}
        \State $row$, $col$ = reprojectToImage(${_q\mathcal{M}_S}[i]$)
        \State $r$ = computeRange(${_q\mathcal{M}_S}[i]$)
        \State $r_{\max}$ = $I_{\max}$[$row$, $col$]
        % \State $r_{\text{bound}}$ = $I_{\text{bound}}$[$row$, $col$]

        \If{$r > r_{\max}$}
            \State // \textit{set maximum depth}
            \State $I_{\max}$[$row$, $col$] = round($r$)
        \EndIf
    \EndFor
    \State $\mathcal{D}_{I}$.append($I_{\max}$)
\EndFor

\State \textbf{Output}: the visibility bounds of all submaps $\mathcal{D}_{I}$.

\end{algorithmic}
\end{algorithm}

\subsubsection{Visibility check}
% visibility check
To approximate the ray-tracing process, we first discretize the range dimension to multiple depth channels. Specifically, the origin image size ($\text{Width} \times \text{Height} \times 1$) is discretized to ($\text{Width} \times \text{Height} \times \text{Depth}$). As shown in \prettyref{fig:broadcast}, the depth channel from one pixel is divided into different cells. The farther the cell is, the more voxels will it contain. Then, we can reproject each submap to the image plane at the query pose $p_q$ according to the specific field of view (FOV) and resolution. Next, for each pixel, we can compute the depth that the point in the pixel can reach. The \textit{visibility boundary} of the pixel is defined by the maximum depth. The cell before the \textit{visibility boundary} is approximated to be passed through by ray-tracing. 

However, such approximation is still affected by the large angle ambiguity and occlusion (i.e., \textit{visibility issues}). Specifically, since a submap is composed of multiple frames and the occlusion is inevitable, the submap will occupy different depths in the same pixel. Similarly, considering the situation with the large incident angle, the same surface may be detected with different depths. These issues will lead to a situation that multiple different depths are occupied in each pixel, even though it is not caused by dynamic points.

% ray-tracing approximation
% To mitigate these issues, we propose a broadcast model [BROADCAST? WHY THIS NAME, PEOPLE WOULD BE CONFUSED]. 
Since the point with a large incident angle can not well approximate the maximum depth range in the submap, we propose the \textit{incident correction}. By computing the incident angle of each point, we can label those points whose incident angle is greater than a threshold as pseudo-occupied points.  We only consider the closest pseudo-occupied point as the \textit{visibility boundary}. Any further points beyond the pseudo-occupied point are not taken into account since the approximation of the ray-tracing may not be accurate. It may preserve more static points but miss some dynamic points out of the boundary. In terms of other common occupied points, we combine the precomputed \textit{visibility boundary} and compute the farthest \textit{visibility boundary} in each pixel. Finally, the visibility boundaries of all submaps can be obtained. The \textit{visibility check} process has been demonstrated in \prettyref{alg:visibility_check}. According to the visibility boundaries, the occupancy probability can be approximately computed as shown in \prettyref{fig:occ_prob}. 

After computing the occupancy probability, we can obtain the final static local map $_q\mathcal{M}_S$ at query poses $p_q$.

\begin{figure}
\centering
% \vspace*{-0.1in}
\includegraphics[width=0.9\linewidth]{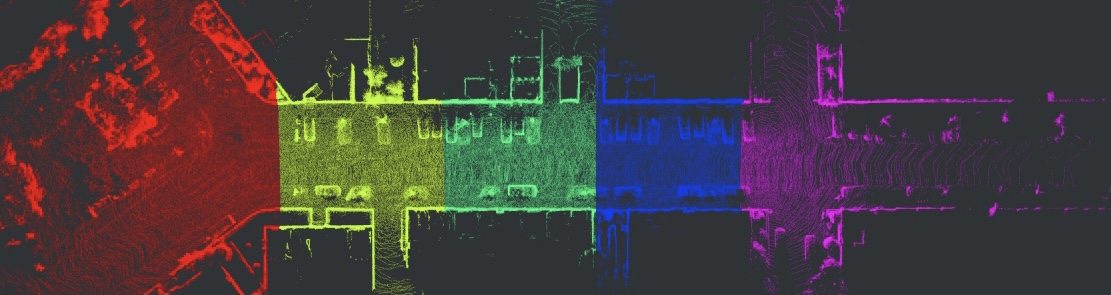}
\caption{The result of the submaps merging in SemanticKITTI. Different colors denote the submap that a voxel belongs to.}
\label{fig:merging}
\vspace*{-0.25in}
\end{figure}

\vspace*{-0.05in}
\subsection{Submaps merging}
However, a single static submap is only sufficient for local scene understanding. Hence, we need to merge multiple static submaps to obtain the global spatial structure.

 The \textit{visibility check} has a better ray-tracing approximation for the points near the submap pose. Therefore, in the merging process, a voxel is marked occupied when it is marked as occupied by the nearest submap, otherwise as free. We visualize the merged result in \prettyref{fig:merging}

\vspace{-0.05in}
\section{Experiments and Results}
\label{sec:exp}
\vspace{-0.05in}
% In this section, we first introduce our experimental setups that include the evaluation metrics, test dataset, simulation benchmark, and comparison approaches. Next, we quantitatively evaluate the our framework with baseline approaches. Then, we conduct the ablation study to validate the efficiency for each component of the framework. Finally, we present our performance in long-sequence and highly-dynamic real-world environments. 

\subsection{Experimental setups}
To measure the quality of the preserved static map after removing dynamic points, we use the \textit{preservation rate} (PR) and \textit{rejection rate} (RR) , proposed by\cite{lim2021erasor}, as our evaluation metrics. Compared to the precision-recall model used in statistics, it is less sensitive to the size of voxelization. 

The metrics are voxel-wise and $0.2$ voxel size is used as the resolution for evaluation. Specifically, the metrics are defined as follows:
\begin{itemize}
    \item PR: $\frac{\# \text{ of preserved static points by the static map}}{\# \text{ of total static points on the raw map}}$
    \item RR: 1 - $\frac{\# \text{ of preserved dynamic points by the static map}}{\# \text{ of total dynamic points on the raw map}}$
\end{itemize}
We also compute the $\text{F}_1$ score which is the harmonic mean of the precision and recall. 
% We choose  

We choose the SemanticKITTI dataset~\cite{geiger2012KITTI,behley2019semantickitti} as our benchmark, which provides the manual label of moving objects and is collected by a vehicle in the urban environment. Although KITTI has been widely used to evaluate SLAM and perception algorithms, the dynamic objects in the dataset do not appear frequently. Hence, the state-of-the-art approaches such as~\cite{lim2021erasor} 
can achieve more than 90\% of PR and RR on the SemanticKITTI dataset. Thus the SemanticKITTI dataset cannot accurately evaluate different algorithms' performance in dynamic object removal.

\begin{figure}
\centering
\includegraphics[width=0.9\linewidth]{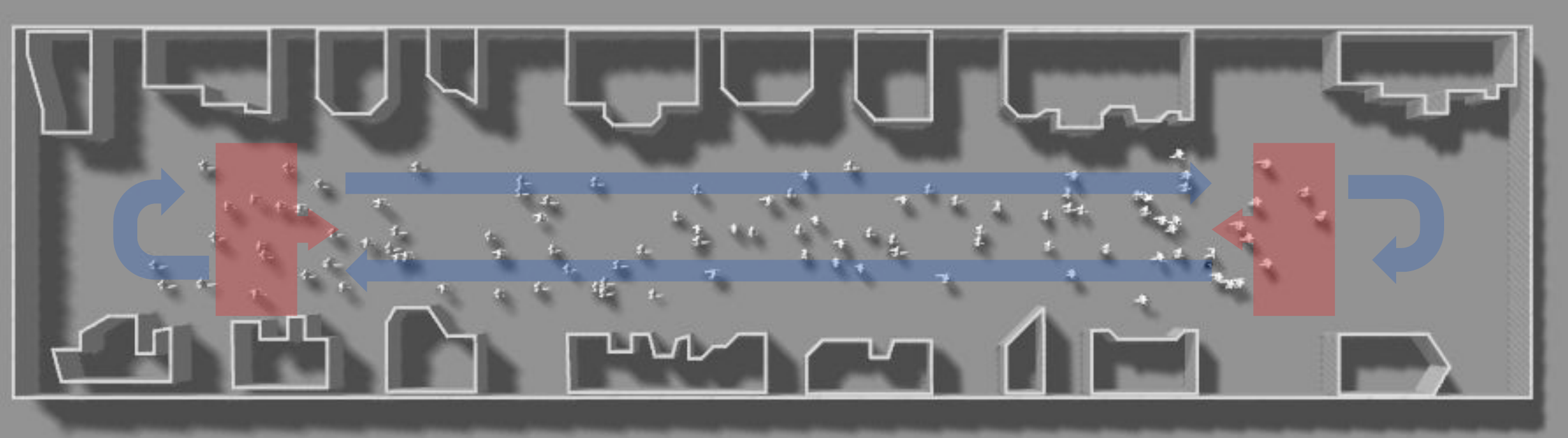}
\caption{Highly dynamic simulation scenario with \SI{70}{m} $\times$ \SI{10}{m}. The red arrow is the moving flow of pedestrians. The blue arrow is the robot trajectory when collecting data.}
\label{fig:simulation}
\vspace*{0in}
\end{figure}

To better evaluate the performance of dynamic object removal, we introduce a simulation environment as demonstrated in \prettyref{fig:simulation} with more dynamic objects, where the Gazebo simulator~\cite{Gazebo_simulator} simulates a mobile robot with a 3D LiDAR and the Menge\cite{Menge} is implemented with the Gazebo environment to simulate the movement of crowded pedestrians. Three scenes with different number of pedestrians (50, 100, 150) are designed. The robot will run two 
rounds to collect data in each scene. Among the collected data in three scenes, dynamic points will take 54.8\%, 58.5\% and 59.3\% of the total points, respectively.

We compare our approach with the state-of-the-art visibility-based approach~\cite{kim2020remove} and the model-dependent approach~\cite{lim2021erasor}. The results show that our approach achieves the state-of-the-art performance on SemanticKITTI over existing approaches~\cite{hornung2013octomap,schauer2018peopleremover}. Note that both approaches are implemented offline. 

\subsection{Quantitative evaluation}
Referring to the setup of~\cite{lim2021erasor}, we only conduct quantitative experiments on the sequences with dynamic objects. As shown in~\prettyref{tab:kitti}, our method achieves comparable performance to that of the state-of-the-art visibility-based approach in semanticKITTI. Note that both approaches ~\cite{kim2020remove, lim2021erasor} are offline in which a pre-built map is needed. In addition, the dynamic points account for less than 1\% of the total number of point clouds in semanticKITTI, and thus it cannot fully test the performance of dynamic object removal.

\begin{table}[]
\vspace{-0.1in}
\centering
\caption{Comparison with state-of-the-art methods on the SemanticKITTI dataset.}
\label{tab:kitti}
\resizebox{0.45\textwidth}{!}{%
\begin{tabular}{llccc}
\hline
Sequence number                  & Method   & PR{[}\%{]} & RR{[}\%{]} & $\text{F}_1$ score \\ \hline
\multicolumn{1}{c}{}   & Removert\cite{kim2020remove} & 86.83      & 90.62      & 0.887    \\
\multicolumn{1}{c}{00} & ERASOR\cite{lim2021erasor}   & \textbf{93.98}      & \textbf{97.08}      &     \textbf{0.955}    \\
                       & Ours     &  90.07          &     91.09       & 0.906      \\ \hline
                       & Removert\cite{kim2020remove} & \textbf{95.82}      & 57.08      & 0.715    \\
\multicolumn{1}{c}{01} & ERASOR\cite{lim2021erasor}   & 91.49      & \textbf{95.38}      &     \textbf{0.934}     \\
                       & Ours     &   87.95         &    87.69        &  0.878        \\ \hline
                       & Removert\cite{kim2020remove} & 83.29      & 88.37      & 0.858    \\
\multicolumn{1}{c}{02} & ERASOR\cite{lim2021erasor}   & 87.73      & \textbf{97.01}      &    \textbf{0.921}      \\
                       & Ours     &    \textbf{88.02}        &   86.10         &   0.871      \\ \hline
                       & Removert\cite{kim2020remove} & 88.17      & 79.98      & 0.839    \\
\multicolumn{1}{c}{05} & ERASOR\cite{lim2021erasor}   & 88.73      & \textbf{98.26}      &    \textbf{0.921}      \\
                       & Ours     &    \textbf{90.17}        &    84.65        &    0.873      \\ \hline
                       & Removert\cite{kim2020remove} & 82.04      & 95.50      & 0.883    \\
\multicolumn{1}{c}{07} & ERASOR\cite{lim2021erasor}   & \textbf{90.62}      & \textbf{99.27}      &   \textbf{0.948}       \\
                       & Ours     &    87.94        &     86.80      &   0.874       \\ \hline
\end{tabular}%
}
\vspace{-0.25in}
\end{table}

As shown in \prettyref{tab:gazebo_exp}, the performance of baseline approaches dramatically degrade in the highly dynamic scenarios that we built in simulation. \cite{lim2021erasor}'s model-dependent approach, relying on the ground fitting to remove dynamic points above the ground, fails in these highly dynamic scenarios where it is not trivial to determine the ground plane. The performance of~\cite{kim2020remove} dramatically decreases in these scenarios due to the \textit{visibility issues}. 

As contrast, our approach achieves more than $90$\% PR and RR in these three scenarios. Moreover, when increasing pedestrian numbers, our approach can still maintain the robust performance. To further clarify the mechanism of our approach, we next conduct a detailed ablation study. 

% \usepackage{graphicx}

% \usepackage{multirow}
% \usepackage{graphicx}
% \usepackage{graphicx}
\begin{table}[]
% \vspace{-0.1in}
\centering
\caption{Comparison experiments in simulation.}
\label{tab:gazebo_exp}
\resizebox{0.45\textwidth}{!}{%
\begin{tabular}{clccc}
\hline
\multicolumn{1}{l}{\# of pedestrians} & Method   & PR{[}\%{]} & RR{[}\%{]} & F1 score \\ \hline
                                      & Removert\cite{kim2020remove} & 69.63      & 64.84      & 0.671    \\
50                                    & ERASOR\cite{lim2021erasor}   & 72.25      & 76.26      & 0.742    \\
\multicolumn{1}{l}{}                  & Ours     & \textbf{95.63}      & \textbf{94.57}      & \textbf{0.951}    \\ \hline
                                      & Removert\cite{kim2020remove} & 66.48      & 57.96      & 0.619    \\
100                                   & ERASOR\cite{lim2021erasor}   & 69.16      & 71.87      & 0.705    \\
\multicolumn{1}{l}{}                  & Ours     & \textbf{95.32}      & \textbf{94.26}      & \textbf{0.948}    \\ \hline
                                      & Removert\cite{kim2020remove} & 68.59      & 51.13      & 0.586    \\
150                                   & ERASOR\cite{lim2021erasor}   & 71.86      & 59.13      & 0.649    \\
\multicolumn{1}{l}{}                  & Ours     & \textbf{95.41}      & \textbf{90.76}      & \textbf{0.930}    \\ \hline
\end{tabular}%
}
\vspace{-0.1in}
\end{table}
\subsection{Ablation study}
We now investigate four different variants of our approach:
\begin{itemize}
    \item \textit{front-end only} method only using our front-end;
    \item \textit{back-end only} method only using our back-end;
    \item \textit{non-visibility check} method which does not use \textit{visibility check} to approximate ${n}_{\text{free}} + {n}_{\text{occ}}$. Instead, we assign the number of the nearest submaps as ${n}_{\text{free}} + {n}_{\text{occ}}$, which assumes all voxels are visible in the submap.  
    \item \textit{non-incident correction} approach that does not consider the pseudo occupied points during the \textit{visibility check}.
\end{itemize}

% \usepackage{multirow}
% \usepackage{graphicx}
\begin{table}[]
\centering
\caption{Ablation study in simulation.}
\label{tab:ablation_study}
\resizebox{0.45\textwidth}{!}{%
\begin{tabular}{clccc}
\hline
\multicolumn{1}{l}{\# of pedestrians} & Method                  & PR{[}\%{]} & RR{[}\%{]} & $\text{F}_1$ score \\ \hline
\multirow{6}{*}{50}                   & Removert\cite{kim2020remove}               & 69.63      & 64.84      & 0.671    \\
                                      & Front-end only          & 80.69      & 62.19      & 0.702    \\
                                      & Back-end only           & \textbf{95.78}      & 92.12      & 0.939    \\
                                      & Non-visibility check    & 74.48      & \textbf{99.66}      & 0.852    \\
                                      & Non-incident correction & 90.96      & 98.92      & 0.948    \\
                                      & Ours                    & 95.63      & 94.57      & \textbf{0.951}    \\ \hline
\multirow{6}{*}{100}                  & Removert\cite{kim2020remove}                & 66.48      & 57.96      & 0.619    \\
                                      & Front-end only          & 78.73      & 57.39      & 0.664    \\
                                      & Back-end only           & 95.06      & 80.85      & 0.874    \\
                                      & Non-visibility check    & 71.82      & \textbf{99.38}      & 0.834    \\
                                      & Non-incident correction & 90.27      & 99.22      & 0.945    \\
                                      & Ours                    & \textbf{95.32}      & 94.26      & \textbf{0.948}    \\ \hline
\multirow{6}{*}{150}                  & Removert\cite{kim2020remove}                & 68.59      & 51.13      & 0.586    \\
                                      & Front-end only          & 76.94      & 61.42      & 0.683    \\
                                      & Back-end only           & 93.83      & 58.32      & 0.719    \\
                                      & Non-visibility check    & 68.12      & \textbf{99.47}      & 0.809    \\
                                      & Non-incident correction & 89.33      & 98.97      & \textbf{0.939}    \\
                                      & Ours                    & \textbf{95.41}      & 90.76      & 0.930    \\ \hline
\end{tabular}%
}
\vspace{-0.25in}
\end{table}

\begin{figure}
\centering
\vspace*{-0.2in}
\includegraphics[width=1.0\linewidth]{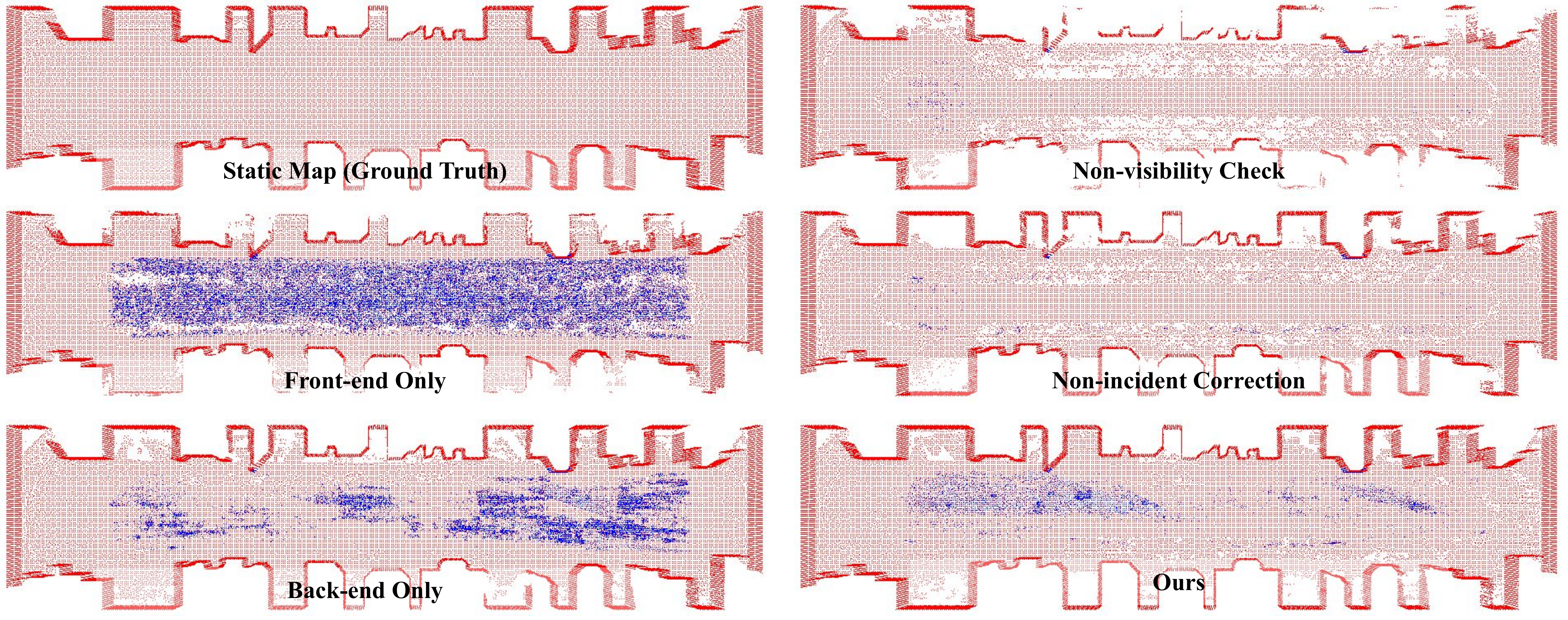}
\caption{Qualitative comparisons for the ablation study. The red points denote the preserved static points; the blue points denote the falsely preserved static points. }
\label{fig:ablation_study}
% \vspace*{-0.2in}
\end{figure}

Similarly, we conduct the ablation study in our simulation environments with different pedestrian numbers. As shown in \prettyref{tab:ablation_study}, the \textit{front-end only} approach outperforms~\cite{kim2020remove}, which mainly benefits from our map-based reverting. The qualitative comparison can be found in \prettyref{fig:revert}. However, the \textit{front-end only} has a limited efficiency in dynamic point removal but can work as the preprocessing in our framework. The \textit{back-end only} method has superior performance when there are a few dynamic objects, but its performance will drop significantly when the number of people increases. The \textit{non-visibility check} method makes ${n}_{\text{free}} + {n}_{\text{occ}}$ larger than the actual value, resulting in a smaller occupancy probability. Although it can remove most of dynamic objects, it also discards some spatial features. The \textit{non-incident correction} and our method have similar performance. Our method with \textit{incident correction} can preserve more static features, but consequentially some dynamic points are also retained. We found that most of the retained dynamic points are close to the ground with a large incident angle. In general, our proposed framework works as a whole with each module closely related. The qualitative comparisons are shown in \prettyref{fig:ablation_study}
.

\begin{figure}
\centering
% \vspace*{-0.1in}
\includegraphics[width=0.8\linewidth]{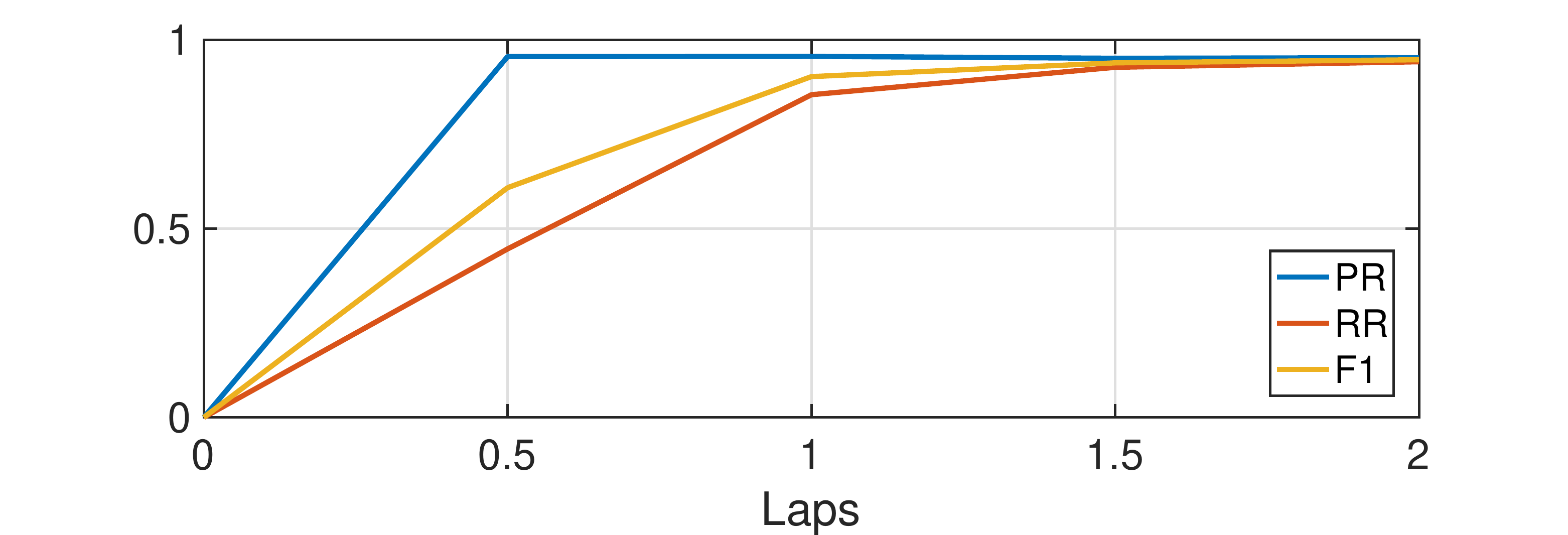}
\caption{Online performance for the simulated robot. }
\label{fig:online_schedule}
\vspace*{-0.1in}
\end{figure}

Additionally, we explore the online performance of the proposed framework. We recorded the online performance when the robot is running in simulation with $100$ pedestrians. As shown in \prettyref{fig:online_schedule}, after the robot runs around the scene once, our approach can retain the spatial structure in the environment and remove most dynamic points. As the number of observations increases, the performance will increase slightly until saturation after going around the scene twice. 

Meanwhile, all the computation can be real-time run on an onboard PC with an Intel i7 8559U, with the processing time of $17.95$ frames per second for the front-end and $11.20$ submaps per second for the back-end.

\vspace*{-0.05in}
\subsection{Real-world experiments}

To more realistically demonstrate the online scene understanding of our approach in the urban environment, we recorded two long sequences of real-world dataset in a shopping mall and an outdoor pedestrian zone. The shopping mall has highly dense pedestrians to be removed. For the outdoor pedestrian zone, dynamic objects become more complex, including not only pedestrians but also bicycles, strollers, dogs, etc. 

\begin{figure}
\centering
% \vspace*{-0.05in}
\includegraphics[width=0.9\linewidth]{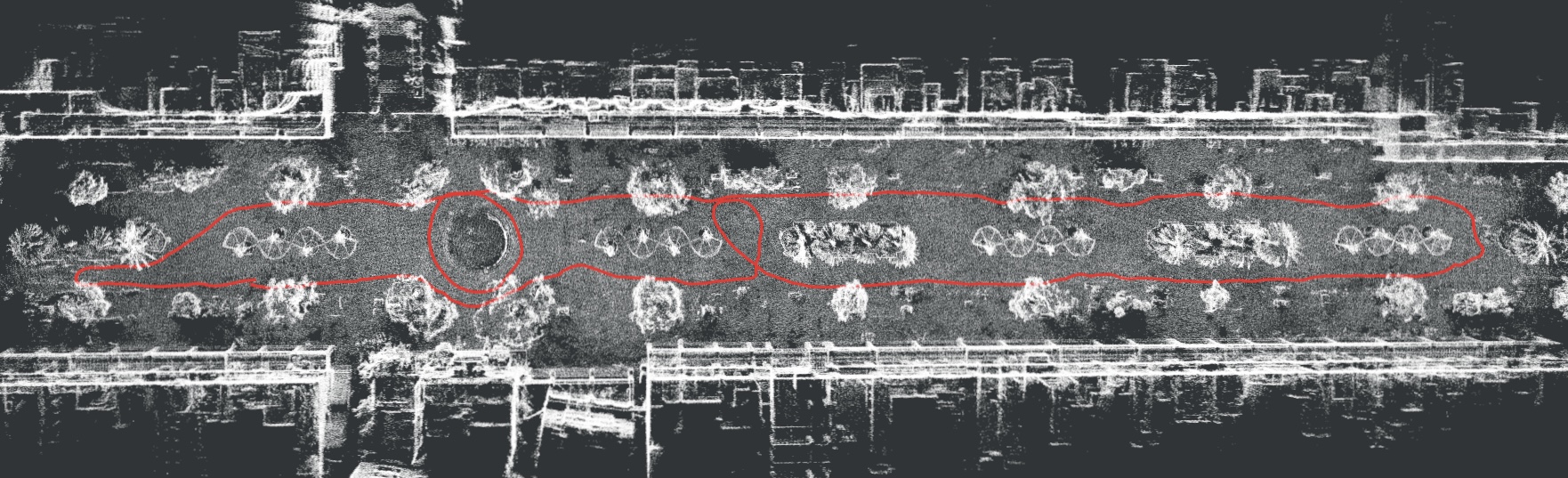}
\caption{Our result in the outdoor pedestrian zone. The white points are the preserved static points. The red line are the LiDAR trajectory. The result is clean without any pedestrians, bicycles, etc.}
\label{fig:xili}
\vspace*{-0.25in}
\end{figure}

Although both scenarios are challenging for online scene understanding, our approach can still online preserve the dense spatial structure of the urban environment as shown in \prettyref{fig:cover_image} and \prettyref{fig:xili}. All real-world experiments can be found in \url{https://sites.google.com/view/dynamicfilter/}.

\vspace{-0.05in}

\section{Conclusion}
\label{sec:conclusion}
\vspace{-0.05in}
In this paper, we proposed an online dynamic objects removal framework that integrates the map-based and visibility-based approaches. The various experiments validate its efficiency in highly dynamic scenarios.

\clearpage
{\small
\bibliographystyle{IEEEtran}
\bibliography{references}
}

\end{document}